\documentclass[runningheads]{llncs}
\usepackage[T1]{fontenc}
\usepackage{graphicx}
\usepackage[utf8]{inputenc}
\usepackage[normalem]{ulem}
\usepackage{multirow}
\usepackage{array}
\usepackage{amsmath}
\usepackage{float}
\usepackage{placeins}

\newcolumntype{x}[1]{>{\centering\arraybackslash\hspace{0pt}}p{#1}}

\begin{document}
\title{DroBoost: An Intelligent Score and Model Boosting Method for Drone Detection\thanks{Supported by OBSS Technology.}}
\titlerunning{DroBoost: Intelligent Score and Model Boosting}
% If the paper title is too long for the running head, you can set
% an abbreviated paper title here
%
\author{Ogulcan Eryuksel \and
Kamil Anil Ozfuttu \and
Fatih Cagatay Akyon \and
Kadir Sahin \and
Efe Buyukborekci \and
Devrim Cavusoglu \and
Sinan Altinuc}
\authorrunning{Eryuksel et al.}
% First names are abbreviated in the running head.
% If there are more than two authors, 'et al.' is used.
%
\institute{OBSS AI\\
OBSS Technology, Universiteler Mah. \\ 
1606. Cad. No:4/1/307 Cyberpark Cyberplaza C blok 3.kat \\
Cankaya/Ankara \\
\url{www.obss.tech} \\ 
\email{\{ogulcan.eryuksel,anil.ozfuttu,fatih.akyon,kadir.sahin,\\
efe.buyukborekci,devrim.cavusoglu,sinan.altinuc\}@obss.tech}}
\maketitle              % typeset the header of the contribution
\begin{abstract}
Drone detection is a challenging object detection task where visibility conditions and quality of the images may be unfavorable, and detections might become difficult due to complex backgrounds, small visible objects, and hard to distinguish objects. Both provide high confidence for drone detections, and eliminating false detections requires efficient algorithms and approaches. Our previous work, which uses YOLOv5, uses both real and synthetic data and a Kalman-based tracker to track the detections and increase their confidence using temporal information. Our current work improves on the previous approach by combining several improvements. We used a more diverse dataset combining multiple sources and combined with synthetic samples chosen from a large synthetic dataset based on the error analysis of the base model. Also, to obtain more resilient confidence scores for objects, we introduced a classification component that discriminates whether the object is a drone or not. Finally, we developed a more advanced scoring algorithm for object tracking that we use to adjust localization confidence. Furthermore, the proposed technique won 1st Place in the Drone vs. Bird Challenge (Workshop on Small-Drone Surveillance, Detection and Counteraction Techniques at ICIAP 2021).

\end{abstract}
\section{Introduction}
Unmanned Aerial Vehicles (UAVs) have been used for some time, and the recent technological and industrial developments have reduced the costs, making them more accessible and abundant in the commercial markets. After a few successful commercial use cases, the demand for drones has boomed. The use cases include surveillance \& security \cite{li2021networked}, photography, delivery \cite{benarbia2021literature}, warehouse operations \cite{wawrla2019applications}, environmental monitoring \cite{fascista2022toward}, etc. With the increase in market capacity, drones can be easily purchased on the Internet at low prices in the present era. Another perspective of drone technology is that they got smaller and better at the assigned tasks as time passed.
Consequently, high reachability brings plenty of opportunities for commercial industry and individuals along with the defense market. On the other hand, as a side-effect of the ease of reachability raises issues about safety, privacy, and security. Thus, misuse of UAVs for illegal activities, invasion of privacy, and violation of regulations need to be addressed.

To address the problems that emerged due to the misuse of drones and to gather the potential solutions under a roof, Drone-vs-Bird Challenge was organized starting from 2017 at the International Workshop on Small-Drone Surveillance, Detection and Counteraction Techniques (WOSDETC) \cite{coluccia2017dvsb}. The challenge aims to assess discrimination between a drone and other (flying) objects at far distances, including similar objects (e.g., birds) from a video dataset. Another challenge apart from detecting drones is that it is not straightforward to make additions to the provided dataset as flying drones often require permission and are restricted in several areas.

% Something here, touching the solutions proposed until today (NN based or other.) (not like related work but as an intro, e.g. some used CNNs, some used newly proposed transformers, earlier works include solutions with a PCA based sensor data for trajectories, ...), and with a few sentences our approach at the end. This will differ from related work in not giving detail about each individual work.

Drone detection, a sub-field of object detection, has been studied especially within the past few years with the challenge's launch. The solutions proposed for drone detection and tracking were various in terms of methodology but mostly based on deep learning in the present era. Some works also include approaches that are not based on deep learning, such as SVMs \cite{srigrarom2020drone}, AdaBoost \cite{gokce2015vbdet}.

In this paper, we propose an approach composed of drone detection and object tracking components. We have taken the methodology of our previous work  \cite{obss2021track} as a baseline and further improved both the detection and tracking components. Firstly, we use a more diverse and balanced dataset that combines several sources. Secondly, we use false negative predictions from synthetically generated drone image data to improve detection performance. We also included various background images with no drones as negative samples to training data where our model created false predictions. We combined real data with these samples to improve cases where base model fails. Thirdly, we used a binary classification algorithm trained on bounding box crops of drone images from training datasets and our YOLOv5 models' false positive predictions on the training data. Model confidence is calculated by the geometric average of YOLOv5 confidence and binary classification confidence. Lastly, we used a specialized scoring algorithm to determine the confidence for the track and adjust the bounding box prediction confidences using this information.

\section{Related Work}

Deep learning-based detection approaches have produced good results for various applications in recent years, including drone detection.

A two-staged detection strategy has been proposed in \cite{sommer2017flying}. First, the authors examined the suitability of different conventional image processing-based object detection techniques, i.e., frame differencing and background subtraction techniques, locally adaptive change detection, and object proposal techniques \cite{muller2017robust}, to extract region candidates in video data from static and moving cameras. In the second stage, a shallow CNN classification network is used to classify each candidate region into drone and clutter categories.

In \cite{coluccia2021drone}, Gagné and Mercier (referred to as the Alexis team) proposed a drone detection approach based on YOLOv3 \cite{redmon2018yolov3} and taking a single RGB frame as input. By integrating an image tiling strategy, this approach can successfully detect small drones in high-resolution images. Alexis Team leveraged the public PyTorch implementation of YOLOv3 with Spatial Pyramid Pooling (YOLOv3-SPP) made available by Ultralytics \cite{yolov3}. In Spatial Pyramid Pooling \cite{he2015spatial}, the input features are processed by pooling layers of varying sizes in parallel and then concatenated to yield fixed-length feature vectors.

Moreover, in \cite{coluccia2021drone} EagleDrone Team proposed a YOLOv5 \cite{yolov5} based drone detection modality with a linear sampling-based data sub-sampling method. They propose using computed loss per image to select the sampling probability. Furthermore, they detect small and low-resolution drones using an ESRGAN-based super-resolution approach.

Recently, authors of \cite{obss2021track} and CARG-UOTTAWA team \cite{coluccia2021drone} proposed YOLOv5 \cite{yolov5} based drone detection modalities utilizing extra training data. Different from the CARG-UOTTAWA team, \cite{obss2021track} proposed a novel track boosting technique to update the confidence scores of the predictions based on a Kalman-based tracker \cite{norfair}. Moreover, \cite{obss2021track} proposes a synthetic data augmentation technique to increase the detection performance further.

\section{Proposed Technique}

The proposed technique is based on our previous work in \cite{obss2021track} which utilizes a YOLOv5 \cite{yolov5} detector and Kalman-based object tracker. Further modifications have been done by improving the synthetic data generation and track boosting stages. Moreover, an extra classification stage is added on top of the detector, and a new scoring algorithm is proposed to increase the tracker performance.

\subsection{Detection Model}
Like in our previous work \cite{obss2021track}, we use YOLOv5 \cite{yolov5} as our object detection algorithm. It has been shown that more computationally intensive detection methods can yield success and, in many cases, better than YOLOv5 \cite{coluccia2021drone}. YOLOv5 is preferred over other algorithms because it yields a good performance as a detector and computational efficiency makes it a better choice for our solution considering the use in real world applications allowing us to achieve real-time performance and advantages in rapid experimentation allowing us to perform more trainings and experiments in any given time.

\subsection{Synthetic Data}
\label{syntheticdata}
\begin{figure}
\includegraphics[width=\textwidth]{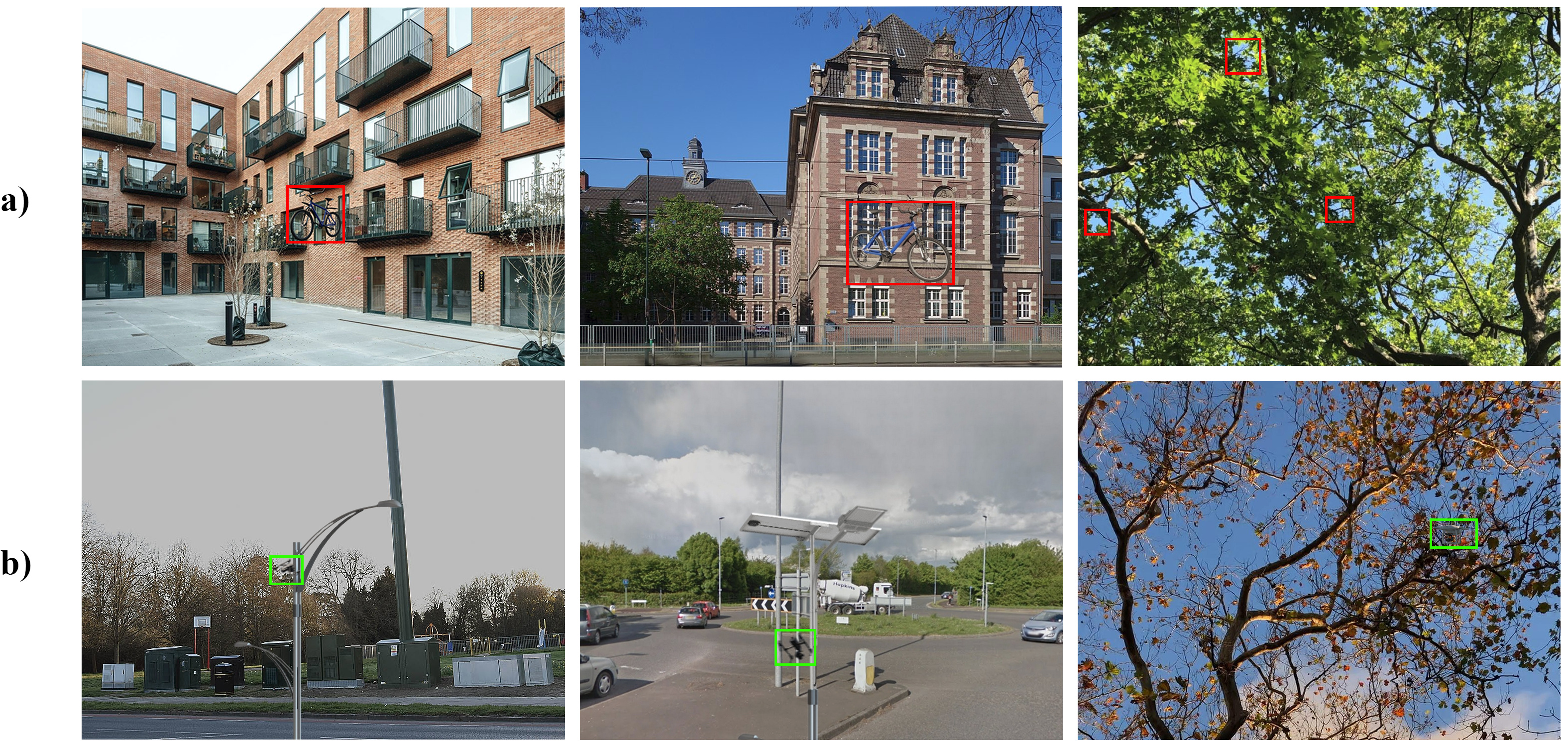}
\caption{Generated synthetic data similar to examples where the real model performs poorly. Synthetic data similar to FP predictions above (a), synthetic data similar to FN predictions below (b).} \label{fig1}
\end{figure}

The use of synthetic data in machine learning training is an increasingly common method for problems where data is difficult to access. The problem of drone detection is one of the areas where it is difficult to obtain data with a large number and variety. In our previous work, we developed a method that produces synthetic data by placing 3D objects on 2D backgrounds and performed some experiments using synthetic and real data. As a result of the experiments, it was seen that the synthetic data alone showed an average performance, and when used with real data, it provided an increase in performance. In this study, various improvements were made to our previous method to overcome the difference between real and synthetic domains, also known as the "domain gap" \cite{DBLP:journals/corr/TobinFRSZA17}.

The use of real-time rendering in synthetic data generation is causing various artifacts around the rendered object. It was observed that these artifacts increased as the object got smaller. Especially when rendered 3D objects on the backgrounds consisting of 2D real images are placed, these artifacts become more evident and negatively affect model performance \cite{ozfuttu2022generating}. For this reason, we developed a method to generate synthetic data using offline rendering techniques.

To increase the performance of the real model, we first identified the cases where the model gave false positive (FP) and false negative (FN) predictions on the test dataset. Then we found similar examples in environments that create FP predictions (such as city and nature environments) and placed 3D models of objects that could be confused with drone objects (such as bicycles and street lamps) on these backgrounds. As a second method, 3D drone objects were placed on backgrounds similar to backgrounds where drones could not be detected, and data similar to FN samples were generated. Also, post-processing effects such as motion blur, filmic noise, and depth of field were applied to generate synthetic data for blending 3D objects with background images.

Synthetic data that we generate according to the cases where the real model makes wrong predictions improves model performance and helps to eliminate FP and FN predictions successfully.

\subsection{Binary Classifier Boosting}

YOLOv5 detectors output category confidence scores in a coupled fashion, together with box regression prediction, from the detection head. Since box regression and classification outputs are trained in a coupled way, it occasionally results in incorrect classification bias, resulting in false positives (FP) having large confidence scores and true positives (TP) having low confidence scores. To overcome this issue, we add a second classification step after YOLOv5 object detector and train a separate image classifier for this purpose. We train this binary image classifier with the cropped TP and FP box predictions from the training set.

For each image in the training set, we perform prediction using the fine-tuned YOLOv5 object detector and acquire $t^{th}$ box prediction $B_t$ and corresponding confidence score $C_t$. Then we calculate intersection over union (IOU) over each predicted box $B_t$ and ground truth box $B^{g}_t$ with a match threshold of 0.3 and assign each prediction as TP or FP. After labeling each box prediction $B_t$, we crop these boxes with a constant margin and create a binary image classification dataset. Ultimately we fine-tune a Vision Transformer (ViT) \cite{dosovitskiy2020image} on this dataset to have a model that predicts whether a given image contains a $drone$ or not. At inference time, we crop the predicted box $B_t$ coming from the object detector and feed into the fine-tuned image classifier to get a classification score $Cl_t$ and calculate the updated confidence score $C^{u}_t$ as:

\begin{equation}
\label{eq:binary_classifier_score}
C^{u}_t =\sqrt{C_t \times Cl_t} 
\end{equation}

Regularizing the detector confidence score with an additional image classification stage decreases the number of FPs and increases the TP scores.

\subsection{Improved Track Boosting and Scoring}

In our previous work \cite{obss2021track}, we had implemented a Kalman-based tracking algorithm to put the images related to a certain drone on the same track. This allowed us to use the temporal information found in images. We had developed a track-boosting algorithm that increased the prediction scores for tracks that are consistent over time. We propose an advanced and more nuanced version of the track-boosting algorithm based on a scoring mechanism. We use the scoring mechanism along with other temporal information to put drones in certainty categories.

The scoring algorithm simply keeps a record of a score for each track. With each new image in the track, the score of the track is modified. As described in equation ~\ref{eq:scoring}, we update the score  ($S_{i}$)  with an amount equal to the difference between a confidence threshold($T_{conf}$) and confidence of the current prediction ($C_i$). We used 0.3 as the threshold in our experiments.

\begin{equation}
\label{eq:scoring}
S_{i} = S_{i-1} + T_{conf} - C_{i} 
\end{equation}

This scoring allows us to keep the negative effect from a low confidence prediction but still allows the system to recover from it if it can provide high confidence predictions in other frames. This performs better than just removing predictions below a certain threshold because low confidence scores do not necessarily mean incorrect predictions, difficult/rare examples may also result in lower confidence scores which can possibly be recovered later with the scoring approach.

As consecutive frames are accumulated in the system, the score usually converges to a positive or a negative value. The longer the track is, the more information is used in scoring, thus allowing better convergence. This allows us to define different categories of certainty for each track. This further improves our previous strategy of averaging the confidence of each frame with maximum confidence in the track by increasing or decreasing confidence values in a more controlled manner.

The calculation of scores for each confidence category is given below. Here ${ S'_{i,j} }$ is the confidence for prediction on ${i}$'th track and ${j}$'th position in the track and ${s_i}$ is the score vector for track ${i}$. 
\begin{itemize}
    \item For tracks that were able to collect higher than a certain score (>25 in our experiments) the tracks are considered highly likely to be a drone. The confidence is $S'_{i,j}= {S_{i,j} * 0.3 + \max(s_i) * 0.7 }$
    \item For tracks with a negative score, we considered them unlikely to be drones. Instead of removing the predictions, we leave them as is since confidence is considerably low: $S'_{i,j}= S_{i,j}$
    \item For tracks that have a relatively low score (<5 in our experiments), if the object has a median velocity below a threshold (0.3 in our experiments), they are considered as stationary objects. The velocity is calculated by the tracker and smoothed out in a window. Confidence for stationary objects are penalized so confidence becomes $S'_{i,j}= S_{i,j} * 0.3 $
    \item Remaining tracks are considered possible drones and confidences are boosted by averaging with the max confidence. $S'_{i,j}= {S_{i,j} * 0.5 + \max(s_i) * 0.5 }$
\end{itemize}

This algorithm can improve its decisions as more information is collected, and it also makes it possible to make decisions on-the-fly without waiting for the end of the track if need be. In this case, the first predictions would be closer to the confidence levels of the detection algorithm but would improve over time. The main advantage of the explained algorithm is that it can make more accurate insights into the predictions by using temporal information. Not only it is helpful to increase the mAP score overall it can also provide some explainability better than using simple threshold values. Expanding the algorithm makes it possible to inject insight gathered from human experience to improve the results without requiring any training. 

\section{Experiments}

\subsection{Datasets}
Different from our previous work \cite{obss2021track}, for training, we have used three real-world datasets in addition to the challenge dataset. Additional real-world datasets are:
\begin{itemize}
    \item Real World Object Detection Dataset For Quadcopter Unmanned Aerial Vehicle \cite{9205392} (named as Real World UAV Dataset for the rest of the paper)
    \item Det-Fly Dataset \cite{Det-Fly}
    \item Multirotor Aerial Vehicle VID (MAV-VID) Dataset \cite{isaac2021unmanned}
\end{itemize}

The Real World UAV Dataset is created from public UAV videos gathered from popular video services. This dataset consists of 51446 images with different image resolutions, ranging from 640x480 to 4k. The dataset contains drone objects from different types, sizes, positions, environments, and lighting conditions.  

The Det-Fly Dataset consists of approximately 13271 images. Image resolutions in the dataset range from 1080p to 4k. The dataset has four environmental backgrounds: sky, urban, field, and mountain. Also, these backgrounds are distributed equally across the dataset. Moreover, drone objects in the images are tiny and challenging. 

The MAV-VID Dataset contains 29500 images for training and 10732 images for validation. Videos in this dataset are captured from other drones, surveillance cameras, and mobile devices. 

\begin{table}[!h]
    % sheet: https://docs.google.com/spreadsheets/d/1GEeLD7PNnug8_nXxZYcZ3eSvN4TEbp7zn-PghNl0wRU/edit?usp=sharing
    \caption{Details of datasets used in the experiments. 
    \newline* The value within brackets '()' represents the number of video samples.}
    \begin{center}
    \begin{tabular}{x{3.75cm}|x{1.5cm}x{1.5cm}x{3cm}x{1.5cm}}
    \hline
    \textbf{Dataset} & \textbf{Type} & \textbf{Split} & \textbf{Total Samples} & \textbf{Used samples}\\
    \hline\hline
    \multirow{2}{*}{\textbf{Drone-vs-Bird}} & \multirow{2}{*}{Video*} & Train & 76818 (63) & 38409\\
     & & Val & 28182 (13) & 1875\\
    \hline
    \multirow{2}{*}{\textbf{Real-world UAV}} & \multirow{2}{*}{Image} & Train & 46299 & 46299\\
     & & Val & 5145 & 1875\\
    \hline
    \multirow{2}{*}{\textbf{Det-Fly}} & \multirow{2}{*}{Image} & Train & 13271 & 11280\\
     & & Val & 1991 & 1875\\
    \hline
    \multirow{2}{*}{\textbf{MAV-VID}} & \multirow{2}{*}{Image} & Train & 29500 & 9834\\
     & & Val & 10732 & 1875\\
    \hline
    \multirow{2}{*}{\textbf{Synthetic \& Negative}} & \multirow{2}{*}{Image} & Train & 100000 & 5000\\
     & & Val & - & -\\
    \hline
    \end{tabular}
    \end{center}
    \label{tab:datasets}
\end{table}

A large amount of data is produced as a result of the synthetic data generation infrastructure mentioned in section \ref{syntheticdata}. In order to find examples that will improve the real model from these data, 3616 false negative samples were extracted from 100k synthetic data. In addition, 1392 false positive samples were extracted from the negative sample pool generated in the same way. Thus, a set of nearly 5000 data consisting of synthetic and negative samples was created.

By combining four real-world datasets, we have increased diversity across drone types, drone sizes, background environments, and lighting conditions. In addition, synthetic and negative image samples are composed to build a synthetic dataset. With the combination of real-world and synthetic datasets, we further increased the diversity of data. The details of the datasets are shown in Table \ref{tab:datasets}.

\subsection{Training Details}
In our experiments, we have used a popular object detection model, YOLOv5. There are lots of available architectures in YOLOv5. We have used YOLOv5m6 model architecture in experiments. We have conducted experiments using only real-world datasets and a combination of real-world datasets and synthetic data. In total four real-world datasets were used for training (Drone vs. Bird, Real World UAV, Det-Fly and MAV-VID). Moreover, we have created a validation dataset that contains 7500 images. The validation dataset is created by composing four real-world datasets. 

In all experiments, the YOLOv5m6 model is fine-tuned for 10 epochs, with 4 batch size and 1920 image size. COCO pre-trained model weights are used in the fine-tuning stage. Also, the best model is chosen using the mAP score computed on the validation dataset. 

\subsection{Results}

To evaluate our proposed methods, we randomly chose 13 videos from Drone vs. Bird dataset. For data sampling, inference and evaluation, our open-source vision framework SAHI (\cite{obss2022sahi}, \cite{obss2021sahi_soft}) is utilized.

YOLOv5 models used in experiments:
\begin{itemize}
    \item \textbf{Real Data Model:} YOLOv5 model trained on real-world datasets.
    \item \textbf{Real+Synthetic Data Model:} YOLOv5 model trained on combination of synthetic dataset and real-world datasets.
\end{itemize}

As seen in Table \ref{tab:eval}, Real+Synthetic Data Model outperforms Real Data Model, always increasing performance by up to 1.2 mAP. Moreover, the classifier boosting technique improves mAP scores by 1.8 and 1.0 for Real Data Model and the Real+Synthetic Data Model, respectively. Finally, applying the track boosting method with classifier boosting achieves 86.3 mAP for Real+Synthetic Data Model. Thus, we marginally increased mAP by 3.3 by combining all of our proposed techniques. 

\begin{table*}[!h]
 \caption{Fine-tuning results for synthetic data augmentation, classifier boosting, and track boosting technique. In the `Technique` column, `YOLOv5` means YOLOv5m6 model is used as a detector, `CB` means model detection confidence scores boosted by using a drone classifier, and `TB` means the proposed track boosting algorithm is applied. `mAP` corresponds to mean average precision at 0.50 intersection over union threshold.}
    \centering
    \begin{tabular}{c|cc}
    Model & Technique & mAP \\
    \hline
    Real+Synthetic Data Model & YOLOv5 + CB + TB & \textbf{86.6}  \\
    Real Data Model & YOLOv5 + CB + TB & 86.3  \\
    \hline
    Real+Synthetic Data Model & YOLOv5 + CB & 85.5  \\
    Real Data Model & YOLOv5 + CB & 85.4  \\
    \hline
    Real+Synthetic Data Model & YOLOv5 & 84.5  \\
    Real Data Model & YOLOv5 & 83.2  \\
    \end{tabular}
    %\vspace*{-3px}
    \label{tab:eval}
    %\vspace*{-15px}
\end{table*}
\FloatBarrier

\section{Conclusion \& Discussion}
Here, we have presented three main approaches that result in improvement of object detection performance: 

\begin{itemize}
    \item Adding synthetic drone image samples and negative background images 
    \item Additional binary classification model
    \item Boosting confidence values using temporal information with a scoring algorithm
\end{itemize}

We observed that using synthetically generated data does not automatically increase detection performance by itself. Adding synthetic data to the model blindly would often perform worse than not including at all. This might result from a domain gap between real data and synthetic data. On the other hand, real-world drone datasets by themselves might fail to provide the necessary generalizability due to the lack in variety of backgrounds and objects. To decrease false negative rate of the model, it was essential to generate synthetic data with complex backgrounds that normally the model performs poorly on. Also, there are various complex objects and backgrounds that can be false positive predictions. Adding various complex objects and backgrounds also provides more generalizability to the model and decrease false positive rates. Using synthetic data with moderate amounts and based on the error analysis of the models outweighs the negative effects of the domain gap.

Objects detection models find objects anywhere on an image, and this requires operating in a very large variety of images and large sample space. However, the predictions of the detection model (whether TP or FP) is a more confined space of images that are drones or at least have drone like features. A binary classification model is shown to be more effective at operating in this confined space, thus separating false predictions from real drone predictions a little better, improving our results.

Using temporal information can improve drone detection performance by boosting the predictions found in the same track \cite{obss2021track}. We showed that it is possible to go further and discriminate confident predictions from non-confident ones based on a scoring algorithm that accumulates confidence scores over time. This can both be used to increase the overall mAP score and introduce more explainability to prediction based on tracks.

%
% ---- Bibliography ----
%
% BibTeX users should specify bibliography style 'splncs04'.
% References will then be sorted and formatted in the correct style.
%
\bibliographystyle{splncs04}
\bibliography{references}
\end{document}